\def\year{2019}\relax
\documentclass[letterpaper]{article} 
\usepackage{aaai19}  
\usepackage{times}  
\usepackage{helvet}  
\usepackage{courier}  
\usepackage{url}  
\usepackage{graphicx}  

\usepackage{xcolor}
\usepackage{soul}
\usepackage[utf8]{inputenc}

\usepackage{booktabs} 
\usepackage[linesnumbered,boxed,ruled]{algorithm2e}
\usepackage{graphicx, subfigure}
\usepackage{float}
\usepackage{enumitem}
\usepackage{comment}%
\usepackage{amsmath,amssymb}
\usepackage{hyperref}       

\usepackage{mdframed}

\usepackage{pifont}

\newtheorem{theorem}{Theorem}

\newtheorem{remark}[theorem]{Remark}

\begin{document}
%
\title{GeniePath: Graph Neural Networks with Adaptive Receptive Paths}
\author{Ziqi Liu$^\dag$, Chaochao Chen$^\dag$, Longfei Li$^\dag$, Jun Zhou$^\dag$  \\
{\bf \Large Xiaolong Li$^\dag$, Le Song$^\dag$$^\ddag$, Yuan Qi$^\dag$} \\
$^\dag$Ant Financial Services Group, China, $^\ddag$Georgia Institute of Technology, Atlanta, USA \\
\{ziqiliu, chaochao.ccc, longyao.llf, jun.zhoujun, xl.li, yuan.qi\}@antfin.com, lsong@cc.gatech.edu
}

\maketitle
\begin{abstract}
We present, GeniePath, a scalable approach for learning adaptive receptive fields 
of neural networks defined on permutation invariant graph data. In GeniePath, we
propose an adaptive path layer consists of two complementary functions designed for breadth and
depth exploration respectively, where the former
learns the importance of different sized neighborhoods, while
the latter extracts and filters signals
aggregated from neighbors of different hops away. Our method works
in both transductive and inductive settings, and extensive
experiments compared with competitive methods
show that our approaches yield state-of-the-art results on large graphs.
\end{abstract}

\section{Introduction}\label{sec:intro}

In this paper, we study the representation learning task involving data
that lie in an irregular domain, i.e. graphs. Many data have the form of a graph, e.g.,
social networks~\cite{perozzi2014deepwalk}, citation networks~\cite{sen2008collective},
biological networks~\cite{zitnik2017predicting}, and transaction
networks~\cite{Liu:2017:PNN:3133956.3138827}. We are interested in graphs
with permutation invariant properties, i.e.,
the ordering of the neighbors for each node is irrelevant to the learning tasks.
This is in opposition to temporal graphs~\cite{kostakos2009temporal}.

Convolutional Neural Networks (CNN) have been proven successful in a diverse range of applications
involving images~\cite{he2016deep} and sequences~\cite{gehring2016convolutional}. 
Recently, interests and efforts have emerged in the literature 
trying to generalize convolutions to graphs~\cite{hammond2011wavelets,defferrard2016convolutional,kipf2016semi,hamilton2017inductive},
which also brings in new challenges.

Unlike image and sequence data that lie in regular domains, graph data are irregular
in nature, making the receptive field of each neuron different for different nodes in the graph.
Assuming a graph $\mathcal{G}=(\mathcal{V}, \mathcal{E})$
with $N$ nodes $i \in \mathcal{V}$, $|\mathcal{E}|$ edges $(i, j) \in \mathcal{E}$,
the sparse adjacency matrix $A \in \mathbb{R}^{N\times N}$,
diagonal node degree matrix $D$ ($D_{ii}=\sum_{ij}A_{ij}$),
and a matrix of node features $X \in \mathbb{R}^{N,P}$. 
Consider the following calculations in typical graph convolutional networks:
$H^{(t+1)} = \sigma \big(\phi(A) H^{(t)} W^{(t)}\big)$ at the $t$-th layer
parameterized by $W^{(t)} \in \mathbb{R}^{K \times K}$, where
$H^{(t)} \in \mathbb{R}^{N \times K}$ denotes the intermediate embeddings of $N$ nodes at the $t$-th layer.
In the case where $\phi(A) = D^{-1}A$ and we ignore the activation function $\sigma$, after $T$ times iterations,
it yields $H^{(T)} = \phi(A)^T H^{(0)} W$\footnote{We collapse $\prod_{i=0}^T W^{(i)}$ as $W$,
  and $H^{(0)}=X \in \mathbb{R}^{N \times P}$.}, with
the $T$-th order transition matrix $\phi(A)^T$~\cite{gagniuc2017markov} as the pre-defined receptive field. That is,
the depth of layers $T$ determines the extent of neighbors to exploit,
and any node $j$ which satisfies $d(i,j) \le T$, where $d$ is the
shortest path distance, contributes to node $i$'s embedding, with
the importance weight pre-defined as $\phi(A)^T_{ij}$. 
Essentially, the receptive field of one target node in
graph domain is equivalent to the subgraph that consists of nodes along paths
to the target node.

Is there a specific path in the graph contributing mostly
to the representation?
Is there an adaptive and automated way of choosing the receptive fields or paths of a graph
convolutional network? It seems that the current literature has not provided such a
solution. For instance, graph convolution neural networks which lie in spectral domain~\cite{bruna2013spectral} heavily
rely on the graph Laplacian matrix~\cite{chung1997spectral}
$L = I - D^{-1/2}A D^{-1/2}$
to define the importance of the neighbors (and
hence receptive field) for each node. Approaches that lie in spatial
domain define convolutions directly on the graph, with receptive field more or
less hand-designed. For instance,
GraphSage~\cite{hamilton2017representation} used the mean or max of a fixed-size neighborhood of each node,
or an LSTM aggregator which needs a pre-selected order of neighboring nodes. 
These pre-defined receptive fields, either based on graph Laplacian in the spectral domain,
or based on uniform operators like mean, max operators in the spatial
domain, thus limit us from discovering meaningful receptive fields from
graph data adaptively. For example, the performance of GCN~\cite{kipf2016semi}
based on graph Laplacian could deteriorate severely if we simply stack more and more layers to
explore deeper and wider receptive fields (or paths), even though
the situation could be alleviated, to some extent, if adding residual nets as extra
add-ons~\cite{hamilton2017representation,velivckovic2017graph}. 

To address the above challenges, 
(1) we first formulate the space of functionals wherein any eligible aggregator functions
should satisfy, for permutation invariant graph data;
(2) we propose adaptive path layer with two complementary components: adaptive breadth and depth functions, 
where the adaptive breadth function can adaptively select a set of significant
important one-hop neighbors, and the adaptive depth function can extract and filter
useful and noisy signals up to long order distance.
(3) experiments on several datasets empirically show that our approaches
are quite competitive, and yield state-of-the-art results on large graphs. Another remarkable result
is that our approach is less sensitive to the depth of propagation layers. 

Intuitively our proposed adaptive path layer guides the breadth and depth exploration
of the receptive fields. As such, we name such adaptively learned receptive
fields as receptive paths.



\section{Graph Convolutional Networks}\label{sec:gcn}

Generalizing convolutions to graphs aims to encode the
nodes with signals lie in the receptive fields. The output
encoded embeddings can be further used in end-to-end supervised
learning~\cite{kipf2016semi} or unsupervised learning tasks
~\cite{perozzi2014deepwalk,grover2016node2vec}.

The approaches lie in spectral domain heavily rely on the
graph Laplacian operator $L = I - D^{-1/2}A D^{-1/2}$~\cite{chung1997spectral}.
The real symmetric positive semidefinite matrix $L$ can be decomposed
into a set of orthonormal eigenvectors which form the graph
Fourier basis $U \in \mathbb{R}^{N\times N}$ such that
$L = U \Lambda U^\top$, where $\Lambda = \mathrm{diag} (\lambda_1, ..., \lambda_n) \in \mathbb{R}^{N\times N}$
is the diagonal matrix with main diagonal entries as ordered
nonnegative eigenvalues. As a result, the convolution operator in spatial
domain can be expressed through the element-wise Hadamard product
in the Fourier domain~\cite{bruna2013spectral}.
The receptive fields in this case depend
on the kernel $U$.

Kipf and Welling~\shortcite{kipf2016semi} further propose GCN to 
design the following approximated localized 1-order spectral
convolutional layer:
\begin{align}\label{eq:gcn}
  H^{(t+1)} = \sigma\big( \tilde{A} H^{(t)} W^{(t)} \big), 
\end{align}
where $\tilde{A}$ is a symmetric normalization of $A$ with self-loops, i.e.
$\tilde{A}=\hat{D}^{-\frac{1}{2}}\hat{A}\hat{D}^{-\frac{1}{2}}$,
$\hat{A}=A+I$, $\hat{D}$ is the diagonal node degree matrix of
$\hat{A}$, $H^{(t)} \in \mathbb{R}^{N,K}$ denotes the $t$-th hidden
layer with $H^{(0)}=X$, $W^{(t)}$ is the layer-specific
parameters, and $\sigma$ denotes the activation functions.

GCN requires the graph Laplacian to normalize the
neighborhoods in advance. This limits the usages of such models in
inductive settings. One trivial solution (GCN-mean) instead is to average the
neighborhoods:
\begin{align}\label{eq:gcn-average}
  H^{(t+1)} = \sigma\big( \tilde{A} H^{(t)} W^{(t)} \big), 
\end{align}
where $\tilde{A} = \hat{D}^{-1}\hat{A}$ is the row-normalized adjacency matrix.

More recently, Hamilton et al.~\shortcite{hamilton2017inductive}
proposed GraphSAGE, a method for computing node representation
in inductive settings. They define a series of aggregator functions
in the following framework:
\begin{align}
  H^{(t+1)} = \sigma\Big( \mathrm{CONCAT}\big(\phi(A, H^{(t)}), H^{(t)}\big) W^{(t)} \Big), 
\end{align}
where $\phi(\cdot)$ is an aggregator function over the graph.
For example, their mean aggregator $\phi(A, H) = D^{-1} A H$ is nearly
equivalent to GCN-mean (Eq.~\ref{eq:gcn-average}), but with additional
residual architectures~\cite{he2016deep}.
They also propose max pooling and LSTM (Long short-term
memory)~\cite{hochreiter1997long} based aggregators. However,
the LSTM aggregator operates on a random permutation of nodes' neighbors
that is not a permutation invariant function with respect to
the ordering of neighborhoods, thus makes the operators hard to understand.

A recent work from~\cite{velivckovic2017graph} proposed 
Graph Attention Networks (GAT) which uses attention mechanisms to parameterize the aggregator function $\phi(A, H; \Theta)$. This method is restricted to learn a direct neighborhood receptive field, which is not as general as our proposed approach which can explore receptive field in both breadth and depth directions. 

Note that, this paper focuses mainly on works related to graph convolutional
networks, but may omit some other types of graph neural networks.

\begin{figure}[t]
        \centering 
        \includegraphics[width=0.34\textwidth,height=0.21\textwidth]{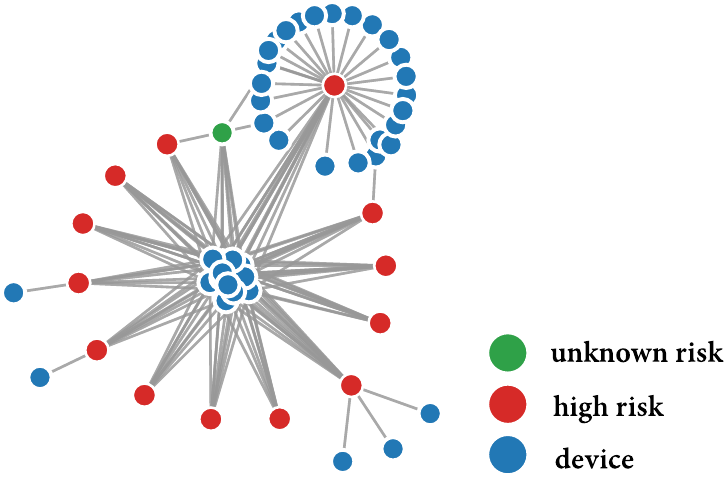}
        \caption{A fraud detection case: accounts with high risk (red nodes), accounts with unknown risk (green nodes), and devices
        (blue nodes). An edge between an account and a device means that the account has logged in via the device during a period. }\label{fig:example}
\end{figure}

\subsection{Discussions}
To summarize, the major effort put on this area is to
design effective aggregator functions that can propagate signals
around the $T$-th order neighborhood for each node.
Few of them try to learn the meaningful paths that direct
the propagation.

Why learning receptive paths is important? The reasons could be twofold:
(1) graphs could be noisy, 
(2) different nodes play different roles, thus it yields different receptive paths. 
For instance, in Figure~\ref{fig:example} we show the graph patterns from
real fraud detection data. This is a bipartite graph where we have two types
of nodes: accounts and devices (e.g. IP proxy). Malicious accounts (red nodes) tend to aggregate together
as the graph tells. However, in reality, normal and malicious accounts could connect
to the same IP proxy. As a result, we cannot simply tell from the graph patterns that
the account labeled in ``green'' is definitely a malicious one. It is important to
verify if this account behave in the similar patterns as other
malicious accounts, i.e. according to the features of each node. Thus,
node features could be additional signals to refine the importance
of neighbors and paths. In addition, we should pay attention to
the number of hops each node can propagate. Obviously,
the nodes at frontier would not aggregate signals from
a hub node, else it would make everything ``over-spread''.

\begin{figure}[t]
        \centering 
        \includegraphics[width=0.28\textwidth,height=0.28\textwidth]{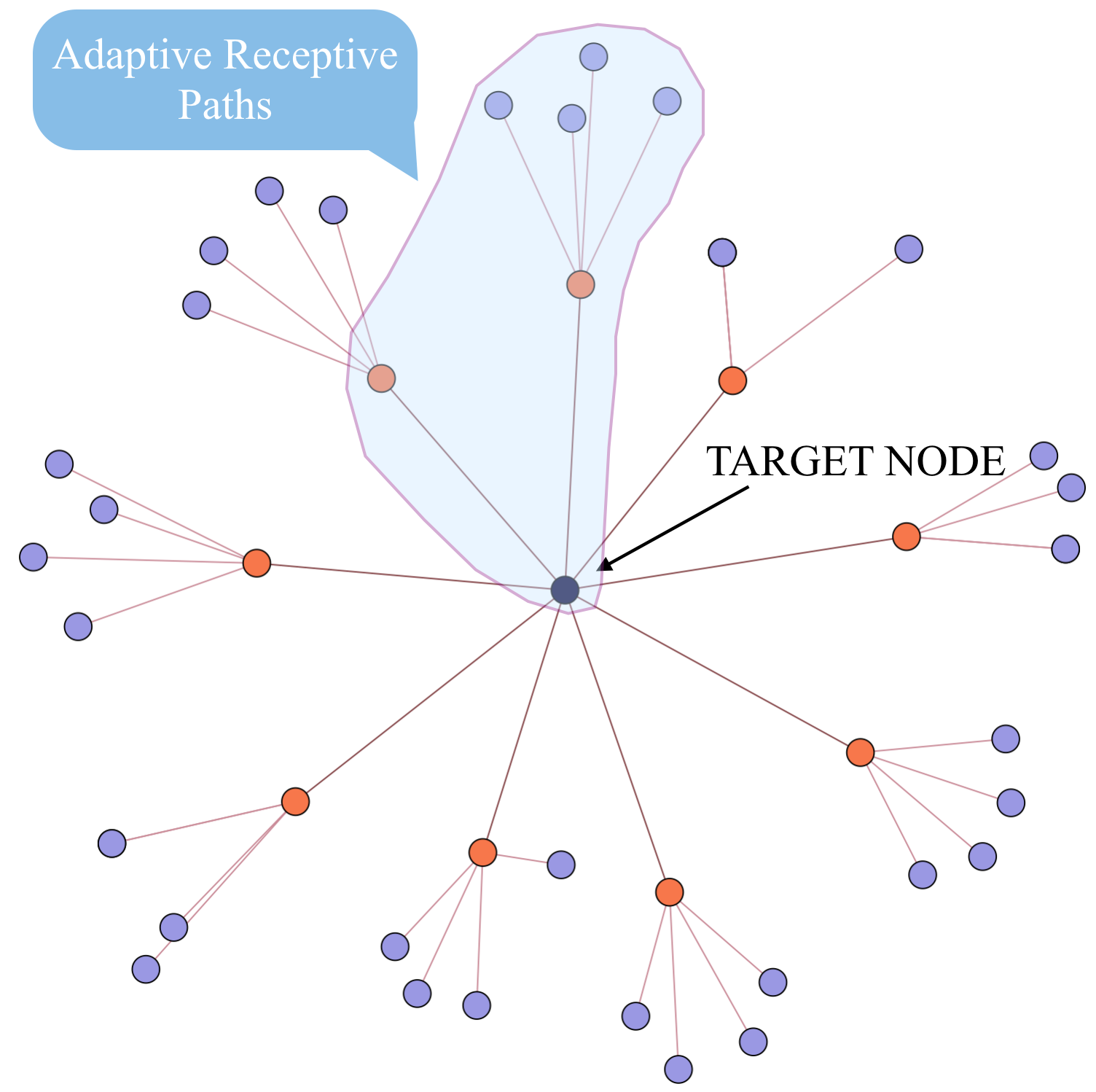}
        \caption{A motivated illustration of meaningful receptive paths (shaded) given all two-hops
          neighbors (red and blue nodes), with the black node as target node.}\label{fig:motivation}
\end{figure}

We demonstrate the idea of learning receptive paths of
graph neural networks in Figure~\ref{fig:motivation}.
Instead of aggregating all the 2-hops
neighbors to calculate the embedding of the target node (black), we
aim to learn meaningful receptive paths (shaded region)
that contribute mostly to the target node.
The meaningful paths can be viewed as a subgraph associated
to the target node. What we need is to do breadth/depth
exploration and filter useful/noisy signals. The breadth exploration
determines which neighbors are important, i.e. leads the direction
to explore, while the depth exploration
determines how many hops of neighbors away are still useful.
Such signal filterings on the subgraphs essentially learn
receptive paths.

\section{Proposed Approaches}
In this section, we first discuss the space of functionals
satisfying the permutation invariant requirement for graph data. Then we
design neural network layers which satisfy such requirement while at the same time have the ability to learn ``receptive paths'' on graph.

\subsection{Permutation Invariant}\label{sec:permutation}
We aim to learn a function $f$ that maps
$\mathcal{G}, \mathcal{H}$, i.e. the graph and associatd (latent) feature spaces, into the range
$\mathcal{Y}$. We have node $i$,
and the 1-order neighborhood associated with $i$, i.e. $\mathcal{N}(i)$, and a set
of (latent) features in vector space $\mathbb{R}^K$ belongs to the neighborhood,
i.e. $\{h_{j} | j \in \mathcal{N}(i)\}$.
In the case of permutation invariant graphs, we 
assume the learning task is independent of the order of neighbors.

Now we require the (aggregator) function $f$ acting on the neighbors
must be invariant to the orders of neighbors
under any random permutation $\sigma$, i.e.
\begin{align}
&f(\{h_1, ..., h_j, ...\}| j \in \mathcal{N}(i)) \\\nonumber
&\quad\quad\quad= f(\{h_{\sigma(1)}, ..., h_{\sigma(j)}, ... \}| j \in \mathcal{N}(i))
\end{align}

\begin{algorithm}[t]
\caption{A generic algorithm of GeniePath.}\label{alg:genie}
\KwIn {Depth $T$, node features $H^{(0)}=X$, adjacency matrix $A$.}
\KwOut{$\Theta$ and $\Phi$}
\While {not converged}
{
  \For{$t=1$ to $T$} { 
    $H^{(tmp)} = \phi(A, H^{(t-1)}; \Theta)$ (\text{breadth function})\\
    $H^{(t)} = \varphi\Big(H^{(tmp)}|\big\{H^{(\tau)}|\tau \in \{t-1,...,0\}\big\};\Phi\Big)$ (\text{depth function})
  }
  Backpropagation based on loss $\mathcal{L}(H^{(T)}, \cdot)$
}
\Return $\Theta$ and $\Phi$
\end{algorithm}

\begin{theorem}[Permutation Invariant]\label{thm:repre}
A function $f$ operating on the neighborhood of $i$ can be a valid function,
i.e. with the assumption of permutation invariant to the neighbors of $i$,
if and only if the $f$ can be decomposed into the form $\rho(\sum_{j \in \mathcal{N}(i)} \phi(h_j))$
with mappings $\phi$ and $\rho$.
\end{theorem}
Proof sketch. It is trivial to show that the sufficiency condition
holds. The necessity condition holds due to the Fundamental
Theorem of Symmetric Functions~\cite{zaheer2017deep}.

\begin{remark}[Associative Property]
If $f$ is permutation invariant with respect to a neighborhood $\mathcal{N}(\cdot)$,
$g \circ f $ is still permutation invariant with respect to the neighborhood
$\mathcal{N}(\cdot)$ if $g$ is independent of the order of $\mathcal{N}(\cdot)$. This allows us to stack the functions in a cascading manner. 
\end{remark}

In the following, we will use the permutation invariant requirement and the associative
property to design propagation layers.  
It is trivial to check that the LSTM aggregator in GraphSAGE is not a valid 
function under this condition, even though it could be possibly an appropriate
aggregator function in temporal graphs.

%

\subsection{Adaptive Path Layer}


Our goal is to learn receptive paths while propagate signals
along the learned paths rather than the pre-defined paths. 
The problem is equivalent to determining a proper subgraph through breadth (which one-hop neighbor
is important) and depth (the importance of neighbors at the $t$-th hop away) expansions
for each node.

To explore the breadth of the receptive paths, we
learn an adaptive breadth function $\phi(A, H^{(t)};\Theta)$ parameterized
by $\Theta$, to aggregate the signals by adaptively assigning different importances to different one-hop neighbors. 
To explore the depth of the receptive paths, we learn an adaptive depth function
$\varphi\Big(h_i^{(t)}|\big\{h_i^{(\tau)}|\tau \in \{t-1,...,0\}\big\};\Phi\Big)$ parameterized by $\Phi$
(shared among all nodes $i \in \mathcal{V}$)
that could further extract and filter the aggregated signals at the $t$-th order distance away
from the anchor node $i$ by modeling the dependencies among aggregated signals at various depths. We
summarize the overall ``GeniePath'' algorithm in Algorithm~\ref{alg:genie}.

The parameterized functions in Algorithm~\ref{alg:genie} are optimized by the customer-defined loss functions.
This may includes supervised learning tasks (e.g. multi-class, multi-label), or unsupervised
tasks like the objective functions defined in~\cite{perozzi2014deepwalk}.




Now it remains to specify the adaptive breadth and depth functions, i.e. $\phi(.)$ and $\varphi(.)$.
We make the adaptive path layer more concrete as follows:
%

\begin{align}
  h^{(\bf tmp)}_i = \tanh \big({W^{(t)}}^\top \sum_{j \in \mathcal{N}(i)\cup \{i\}} \alpha(h_i^{(t)}, h_j^{(t)}) \cdot h_j^{(t)}\big) \label{eq:attention_layer}
\end{align}
then 
\begin{equation}
  \begin{array}{ll}
    i_i = \sigma({W_i^{(t)}}^\top\ h_i^{(\bf tmp)}), & f_i = \sigma({W_f^{(t)}}^\top\ h_i^{(\bf tmp)}) \cr
    o_i = \sigma({W_o^{(t)}}^\top\ h_i^{(\bf tmp)}), & \tilde{C} = \tanh({W_c^{(t)}}^\top\ h_i^{(\bf tmp)}) \cr
    C_i^{(t+1)} = f_i \odot C_i^{(t)} + i_i \odot \tilde{C}, & 
  \end{array} \nonumber
\end{equation}
and finally, 
\begin{align}
  h_i^{(t+1)} = o_i \odot \tanh(C_i^{(t+1)}).
\end{align}

The first equation (Eq.~\eqref{eq:attention_layer}) corresponds to $\phi(.)$ and the rest gated units correspond to $\varphi(.)$.
We maintain for each node $i$ a memory $C_i$ (initialized as $C_i^{(0)} \gets \vec{0} \in \mathbb{R}^K$)
, and gets updated as the receptive paths being explored $t=0,1,...,T$.
At the $(t+1)$-th layer, i.e. while we are exploring the neighborhood $\{j | d(i,j) \leq t+1\}$, we have 
the following two complementary functions.

{\bfseries Adaptive breadth function.}
Eq.~\eqref{eq:attention_layer} assigns the
importance of any one-hop neighbors' embedding $h_j^{(t)}$ by the parameterized generalized linear attention operator $\alpha(\cdot,\cdot)$ 
as follows
\begin{align}\label{eq:att-mech}
\alpha(x,y) &= \mathrm{softmax}_y \big(v^\top \tanh(W_s^\top\  x + W_d^\top\  y)\big) ,
\end{align}
where $\mathrm{softmax}_y f(\cdot,y) = \frac{\exp f(\cdot,y)}{\sum_{y'} \exp f(\cdot,y')}$.

{\bfseries Adaptive depth function.} The gated unit $i_i$ with sigmoid output (in the range $(0,1)$)
is used to extract newly useful signals from $\tilde{C}$, and be added to the memory
as $i_i \odot \tilde{C}$. The
gated unit $f_i$ is used to filter useless signals from the old memory given the newly observed neighborhood $\{j | d(i,j) \leq t+1\}$ by $f_i \odot C_i^{(t)}$.
As such, we are able to filter the memory for each node $i$ as $C_i^{(t+1)}$ while we extend
the depth of the neighborhood.
Finally, with the output gated unit $o_i$ and the latest memory $C_i^{(t+1)}$, we can
output node $i$'s embedding at $(t+1)$-th layer as $h_i^{(t+1)}$.

\begin{figure}[t]
        \centering 
        \includegraphics[width=0.43\textwidth,height=0.16\textwidth]{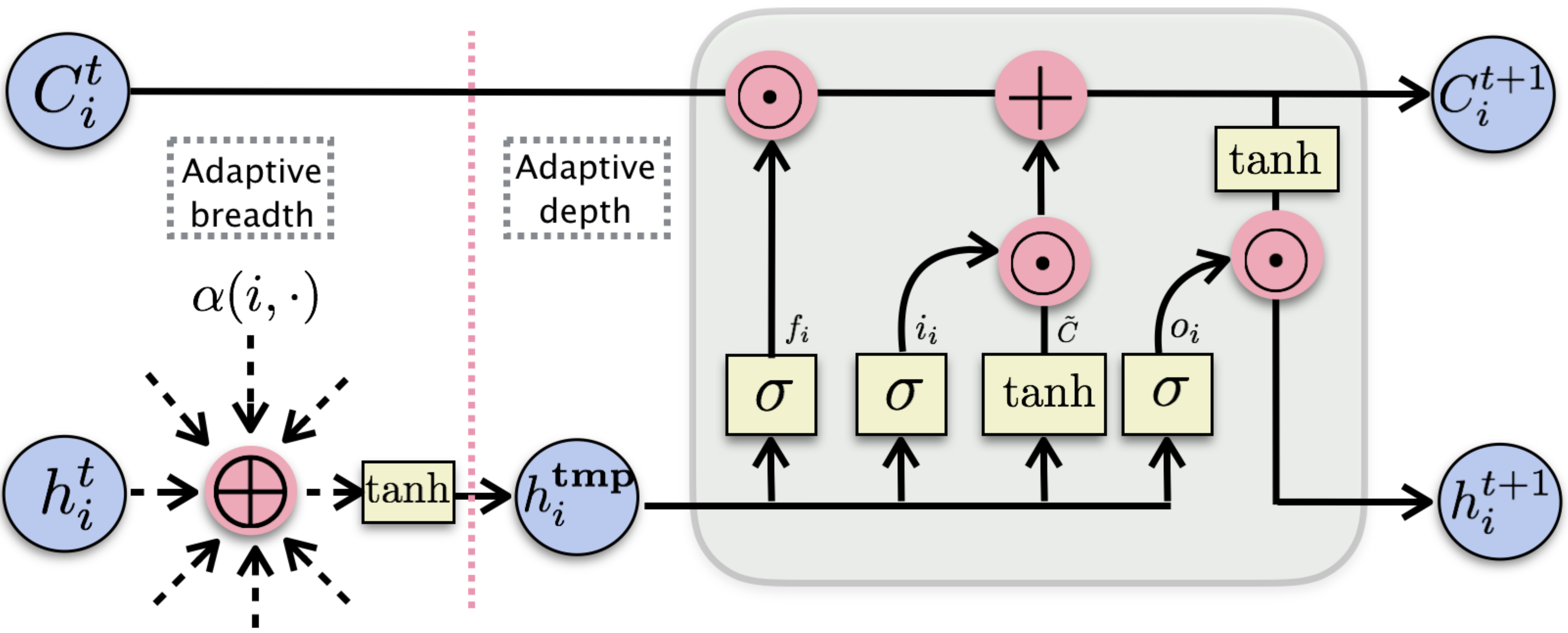}
    \caption{A demonstration for the architecture of GeniePath. Symbol $\bigoplus$ denotes the
    operator $\sum_{j \in \mathcal{N}(i)\cup \{i\}} \alpha(h_i^{(t)}, h_j^{(t)}) \cdot h_j^{(t)}$.}\label{fig:arch}
\end{figure}

The overall architecture of adaptive path layer is illustrated in Figure~\ref{fig:arch}.
We let $h_i^{(0)} = W_x^\top X_i$ with $X_i \in \mathbb{R}^{P}$ as node $i$'s feature.
The parameters to be optimized are weight matrix $W_x \in \mathbb{R}^{P,K}$,
$\Theta = \{W, W_s, W_d \in \mathbb{R}^{K,K}, v \in \mathbb{R}^K\}$, and $\Phi = \{W_i, W_f, W_o, W_c \in \mathbb{R}^{K,K}\}$
that are extremely compact (at the same scale as other graph neural networks).

Note that the adaptive path layer with only adaptive breadth function $\phi(.)$ reduces to
the proposal of GAT, but with stronger nonlinear representation capacities. Our generalized linear
attention can assign symmetric importance with constraint $W_s = W_d$. We do not limit $\varphi(\cdot)$
as a LSTM-like function but will report this architecture in experiments.

{\bfseries A variant.} Next, we propose a variant called ``GeniePath-lazy''
that postpones the evaluations of the adaptive depth function $\varphi(.)$ at each layer.
We rather propagate signals up to $T$-th order distance by merely stacking adaptive breadth functions.
Given those hidden units $\{h_i^{(0)}, ..., h_i^{(t)}, ..., h_i^{(T)}\}$, we add adaptive depth function $\varphi(.)$ on top of them to further
extract and filter the signals at various depths.
We initialize $\mu_i^{(0)} = W_x^\top X_i$, and feed $\mu_i^{(T)}$ to the final
loss functions. Formally given $\{h_i^{(t)}\}$ we have:
{\small
\begin{align}
&i_i = \sigma\big({W_i^{(t)}}^\top\ \mathrm{CONCAT}(h_i^{(t)}, \mu_i^{(t)})\big),\\\nonumber
&f_i = \sigma({W_f^{(t)}}^\top\ \mathrm{CONCAT}\big(h_i^{(t)}, \mu_i^{(t)})\big),\\\nonumber
&o_i = \sigma({W_o^{(t)}}^\top\ \mathrm{CONCAT}\big(h_i^{(t)}, \mu_i^{(t)})\big),\\\nonumber
&\tilde{C} = \tanh({W_c^{(t)}}^\top\ \mathrm{CONCAT}\big(h_i^{(t)}, \mu_i^{(t)})\big),\\\nonumber
&C_i^{(t+1)} = f_i \odot C_i^{(t)} + i_i \odot \tilde{C},\\\nonumber
&\mu_i^{(t+1)} = o_i \odot \tanh(C_i^{(t+1)}).
\end{align}
}

{\bfseries Permutation invariant.} Note that the adaptive breadth function $\phi(\cdot)$ in our case operates on a linear transformation
of neighbors, thus satisfies the permutation invariant requirement stated in Theorem~\ref{thm:repre}.
The function $\varphi(\cdot)$ operates on layers at each depth
is independent of the order of any 1-step neighborhood, thus the composition of $\varphi \circ \phi$ is
still permutation invariant.

\subsection{Efficient Numerical Computation}
Our algorithms could be easily formulated in terms of linear algebra,
thus could leverage the power of Intel MKL~\cite{intel2007intel} 
on CPUs or cuBLAS~\cite{nvidia2008cublas} on GPUs. The major challenge
is to numerically efficient calculate Eq.~\eqref{eq:attention_layer}. For example,
GAT~\cite{velivckovic2017graph} in its first version performs attention over all
nodes with masking applied by way of an additive bias of $-\infty$ to the masked
entries' coefficients before apply the softmax, which results into $\mathcal{O}(N^2)$ in terms of computation and storage.

\begin{table*}
  \centering
  \caption{Dataset summary.}
  \label{tb:data}
  \begin{tabular}{ccccccc}
    Dateset & V & E & \# Classes & \# Features & Label rate (train / test)\\
    \midrule
    Pubmed &          $19,717$ & $88,651$  & $3$  & $500$ & 0.3\% / 5.07\% \\
    BlogCatalog$^1$&  $10,312$ & $333,983$ & $39$ & $0$ & 50\% / 40\% \\
    BlogCatalog$^2$&  $10,312$ & $333,983$ & $39$ & $128$ & 50\% / 40\% \\
    PPI &             $56,944$ & $818,716$  & $121$ & $50$ & 78.8\% / 9.7\%\\
    Alipay&  $981,748$ & $2,308,614$  & $2$ & $4,000$ & 20.5\% / 3.2\% \\
  \bottomrule
\end{tabular}
\end{table*}

Instead, all we really need
is the attention entries in the same scale as adjacency matrix $A$,
i.e. $|\mathcal{E}|$.
Our trick is to build two auxiliary sparse matrices $L \in \{0,1\}^{|\mathcal{E}|,N}$ and
$R \in \{0,1\}^{|\mathcal{E}|,N}$ both with $|\mathcal{E}|$ entries.
For the $i$-th row of $L$ and $R$, we have $L_{ii'}=1$, $R_{ij'}=1$, and $(i',j')$ corresponds
to an edge of the graph.
After that, we can do $L H W_s + R H W_d$ for the transformations on all the edges
in case of generalized linear form in Eq.~\eqref{eq:att-mech}.
As a result, our algorithm complexity and storage is still in linear with $\mathcal{O}(|\mathcal{E}|)$
as other type of GNNs.

\section{Experiments}
In this section, we first discuss the experimental results of our approach
evaluated on various types of graphs compared with strong baselines.
We then study its abilities of learning adaptive depths.
Finally we give qualitative analyses on the learned
paths compared with graph Laplacian.

\subsection{Datasets}

{\bfseries Transductive setting.} 

The \emph{Pubmed}~\cite{sen2008collective} is a type of citation networks, where nodes correspond to documents and
edges to undirected citations. 
The classes are exclusive, so we treat the problem as a multi-class
classification problem.
We use the exact preprocessed
data from Kipf and Welling~\shortcite{kipf2016semi}.

The \emph{BlogCatalog$^1$}~\cite{zafarani2009social} is a type of social networks, where nodes correspond to bloggers
listed on BlogCatalog websites, and the edges to the social relationships of those
bloggers. 
We treat the problem as a multi-label classification problem.
Different from other datasets, the BlogCatalog$^1$ has no explicit features available, as a result, we
encode node ids as one-hot features for each node, i.e. $10,312$ dimensional features.
We further build dataset \emph{BlogCatalog$^2$} with $128$ dimensional features decomposed
by SVD on the adjacency matrix $A$.


The \emph{Alipay} dataset~\cite{Liu:2017:PNN:3133956.3138827} is a type of Account-Device
Network, built for detecting malicious accounts
in the online cashless payment system at Alipay. The nodes correspond to users' accounts and
devices logged in by those accounts. The edges correspond to the login relationships between
accounts and devices during a time period. Node features are counts of login behaviors discretized
into hours and account profiles. There are $2$ classes in this dataset, i.e. malicious
accounts and normal accounts. The Alipay dataset is random sampled during one week.
The dataset consists of $82,246$ disjoint subgraphs.

{\bfseries Inductive setting.} 

The \emph{PPI}~\cite{hamilton2017inductive} is a type of protein-protein interaction networks, which consists of 24 subgraphs
with each corresponds to a human tissue~\cite{zitnik2017predicting}. The node features are extracted by positional
gene sets, motif gene sets and immunological signatures. There are 121 classes for each node 
from gene ontology. 
Each node could have multiple labels, then results into a multi-label classification problem. We use the exact
preprocessed data provided by Hamilton et al.~\shortcite{hamilton2017inductive}. There are 20 graphs for training,
2 for validation and 2 for testing.

We summarize the statistics of all the datasets in Table~\ref{tb:data}.

\begin{table*}
\parbox{.45\linewidth}{
\centering
  \caption{Summary of testing results on Pubmed, BlogCatalog and Alipay
    in the transductive setting. In accordance with former benchmarks, we
    report accuracy for Pubmed, Macro-F1 for BlogCatalog, and F1 for Alipay.}
  \label{tb:baseline_transductive}
\begin{tabular}{ccccc}
    \multicolumn{5}{c}{Transductive} \\
    \toprule
    Methods & Pubmed & BlogCatalog$^1$ & BlogCatalog$^2$ & Alipay \\
    \midrule
    MLP & 71.4\% & - & 0.134 & 0.741 \\
    node2vec & 65.3\% & 0.136 & 0.136 &  - \\
    Chebyshev & 74.4\% & 0.160 & 0.166 &  0.784 \\
    GCN & \textbf{79.0\%} & 0.171 & 0.174  & 0.796 \\
    GraphSAGE$^*$ & 78.8\% & 0.175 & 0.175  & 0.798 \\
    GAT & 78.5\% & \textbf{0.201} & 0.197 & 0.811 \\
    \midrule
    GeniePath$^*$ & 78.5\% & 0.195 & \textbf{0.202} & \textbf{0.826}\\
  \bottomrule
\end{tabular}
}
\hfill
\parbox{.45\linewidth}{
\centering
  \caption{Summary of testing Micro-F1 results on PPI in the inductive setting.}
  \label{tb:baseline_inductive}
\begin{tabular}{cc}
    \multicolumn{2}{c}{Inductive} \\
    \toprule
    Methods & PPI \\
    \midrule
    MLP & 0.422 \\
    GCN-mean & 0.71 \\
    GraphSAGE$^*$ & 0.768\\
    GAT & 0.81 \\
    \midrule
    GeniePath$^*$ & \textbf{0.979} \\
  \bottomrule
\end{tabular}
}
\end{table*}

\begin{table}
\centering
  \caption{A comparison of GAT, GeniePath, and additional residual
  ``skip connection'' on PPI.}
  \label{tb:ablation}
\begin{tabular}{cc}
    \toprule
    Methods & PPI \\
    \midrule
    GAT & 0.81 \\
    \emph{GAT-residual} & 0.914 \\
    \midrule
    GeniePath & 0.952 \\
    GeniePath-lazy & 0.979 \\
    \emph{GeniePath-lazy-residual} & \textbf{0.985} \\
  \bottomrule
\end{tabular}
\end{table}

\subsection{Experimental Settings}

\subsubsection{Comparison Approaches}
We compare our methods with several strong baselines.

(\textbf{1}) MLP (multilayer perceptron), utilizes only the node features but not the structures of the graph.
(\textbf{2}) node2vec~\cite{grover2016node2vec}.
  Note that, this type of methods built on top of lookup embeddings cannot work
  on problems with multiple graphs, i.e. on datasets Alipay and PPI,
  because without any further constraints, the entire embedding space cannot guarantee to be
  consistent during training~\cite{hamilton2017inductive}.
(\textbf{3}) Chebyshev~\cite{defferrard2016convolutional}, which approximates the graph spectral
convolutions by a truncated expansion in terms of Chebyshev polynomials up to $T$-th order.
(\textbf{4}) GCN~\cite{kipf2016semi}, which is defined in Eq.~\eqref{eq:gcn}. Same as Chebyshev, it
  works only in the transductive setting. However, if we just use the normalization
  formulated in Eq.~\eqref{eq:gcn-average}, 
  it can work in an inductive setting. We denote this variant of GCN as GCN-mean.
  (\textbf{5}) GraphSAGE~\cite{hamilton2017inductive}, which consists of a group of pooling
  operators and skip connection architectures as discussed in section~\ref{sec:gcn}.
  We will report the best results of GraphSAGEs with
different pooling strategies as GraphSAGE$^*$.
(\textbf{6}) Graph Attention Networks (GAT)~\cite{velivckovic2017graph} 
is similar to a reduced version of our approach with only adaptive breadth function.
This can help us understand the usefulness of adaptive breadth and depth function.



We pick the better results from GeniePath or GeniePath-lazy as GeniePath$^*$.
We found out that the residual architecture (skip connections)~\cite{he2016identity} are 
useful for stacking deeper layers for various approaches and will
report the approaches with suffix ``-residual''
to distinguish the contributions between graph convolution operators and
skip connection architecture.

\subsubsection{Experimental Setups}
In our experiments, we implement our algorithms in TensorFlow~\cite{Abadi2016TensorFlow}
with the Adam optimizer~\cite{kingma2014adam}. For all the graph
convolutional network-style approaches, we set the
hyperparameters include the dimension of embeddings or
hidden units, the depth of hidden layers and learning rate to be same.
For node2vec, we
tune the return parameter $p$ and in-out parameter $q$ by grid search. Note that setting $p=q=1$
is equivalent to DeepWalk~\cite{perozzi2014deepwalk}.
We sample 10
walk-paths with walk-length as 80 for each node in the graph.
Additionally, we tune the penalty of $\ell_2$ regularizers for different approaches.

{\bfseries Transductive setting.} In transductive settings, we allow all the algorithms
to access to the whole graph, i.e. all the edges among the nodes, and all of the node features.

For pubmed, we set the number of hidden units as 16 with 2 hidden layers.
For BlogCatalog$^1$ and BlogCatalog$^2$, we set the number of hidden
units as 128 with 3 hidden layers.
For Alipay, we set the number of hidden units as 16 with 7 hidden layers.

{\bfseries Inductive setting.} In inductive settings, all the nodes in test are \emph{completely
unobserved} during the training procedure. For PPI, we set the number of hidden units as 256 with 3 hidden layers.


\begin{figure*}[th!]
	\centering 
	\includegraphics[width=0.35\textwidth,height=0.3\textwidth]{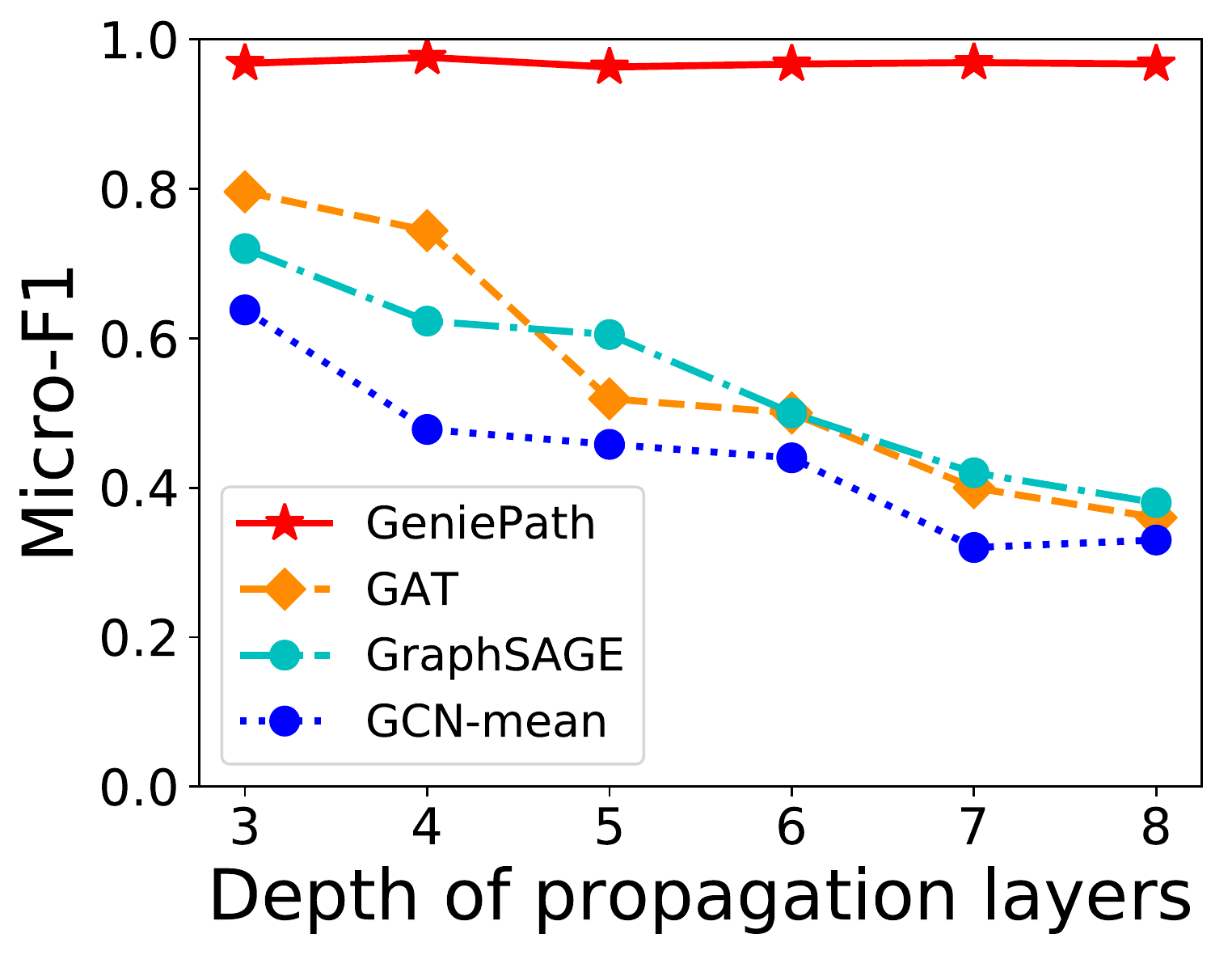}
    \includegraphics[width=0.35\textwidth,height=0.3\textwidth]{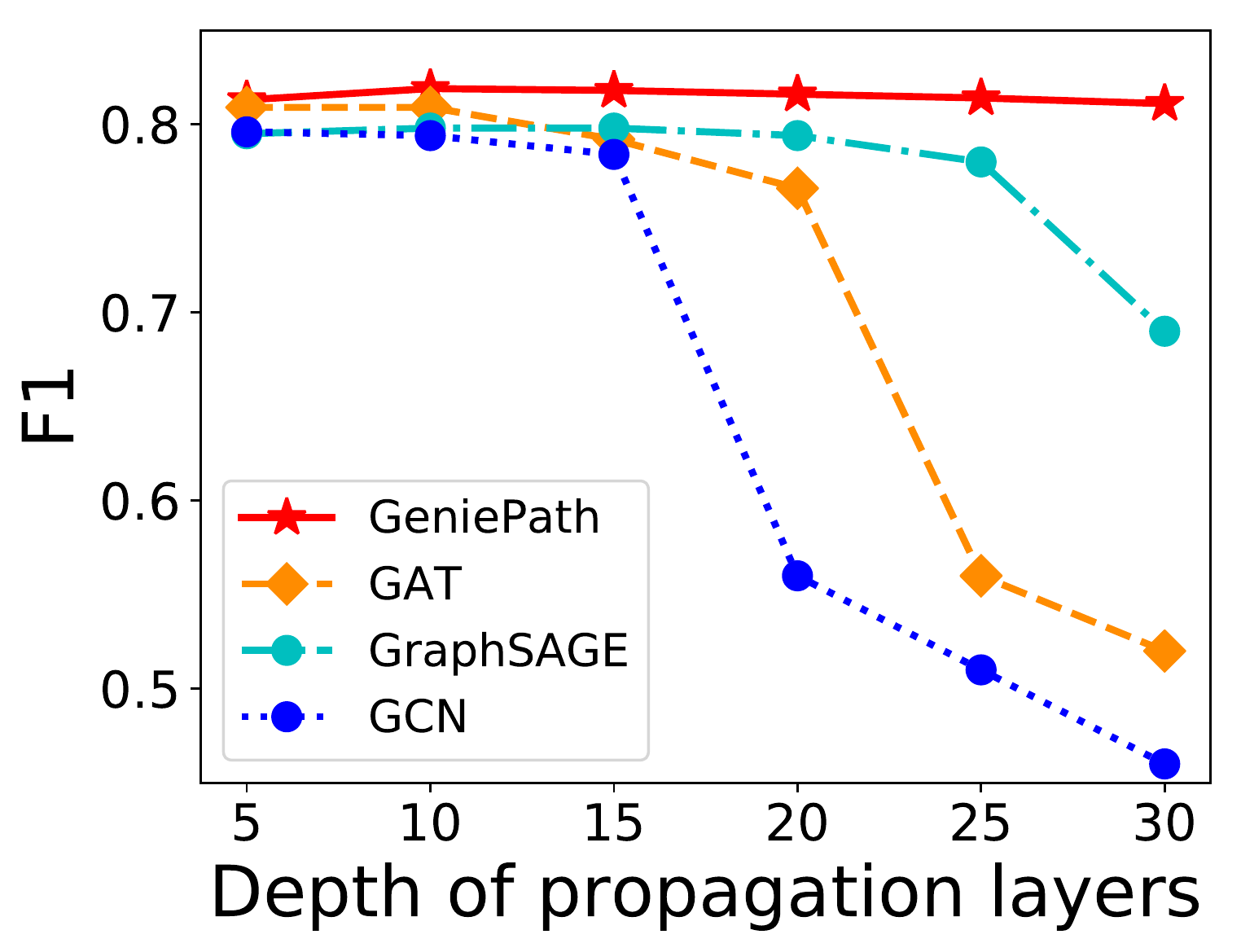}
    \caption{The classification measures with respect to the depths of propagation layers: PPI~(\textbf{left}), Alipay~(\textbf{right}).}\label{fig:prop_depth}
\end{figure*}

\begin{figure*}[th!]
	\centering
	\includegraphics[width=0.45\textwidth,height=0.6\textwidth]{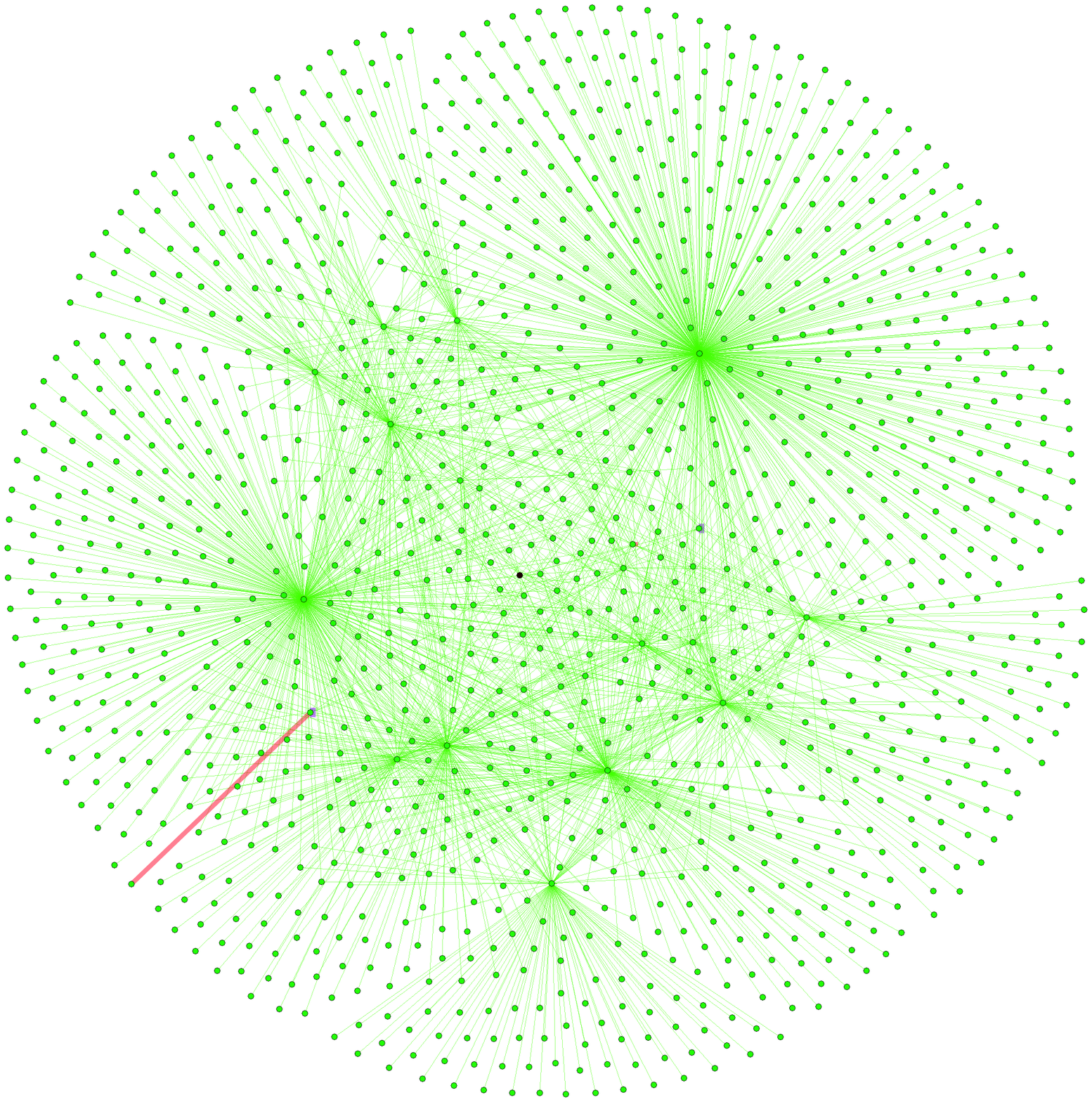}
    \includegraphics[width=0.45\textwidth,height=0.6\textwidth]{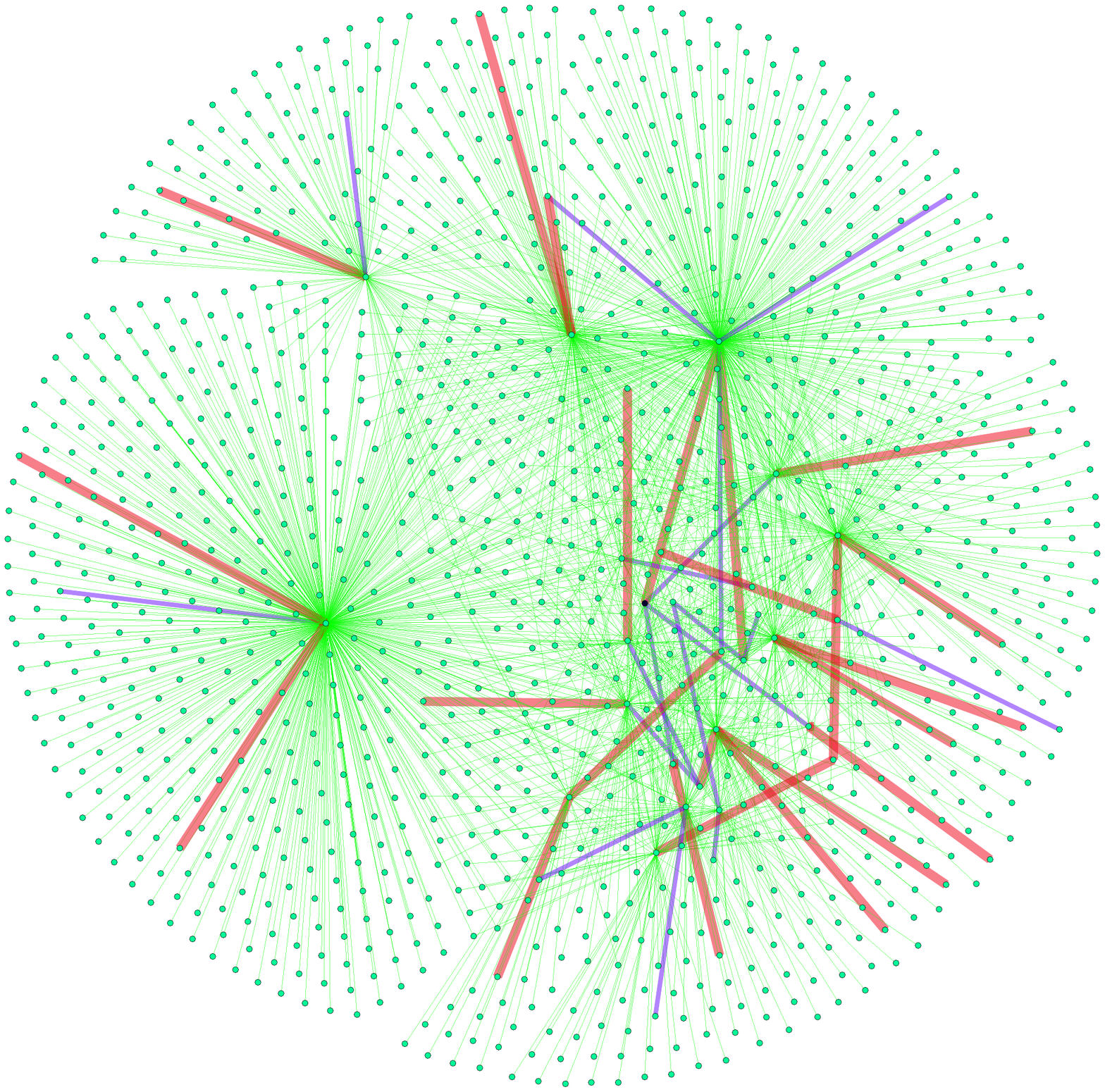}
    \caption{Graph Laplacian (\textbf{left}) v.s. Estimated receptive paths (\textbf{right}) with respect to the \emph{black node} on
    PPI dataset: we retain all the paths to the black node in 2-hops that involved in the propagation,
    with edge thickness denotes the importance $w_{i,j} = \alpha(h_i,h_j)$ of edge $(i,j)$ estimated in the first adaptive path layer.
    We also discretize the importance of each edge into 3 levels: Green ($w_{i,j}<0.1$), Blue ($0.1 \le w_{i,j} <0.2$),
    and Red ($w_{i,j} \ge 0.2$).}\label{fig:qualitative}
\end{figure*}

\subsection{Classification}
We report the comparison results of transductive settings in Table~\ref{tb:baseline_transductive}.
For Pubmed, there are only 60 labels available for training.
We found out that GCN works the best by stacking exactly 2 hidden convolutional layers. Stacking
2 more convolutional layers will deteriorate the performance severely.
Fortunately, we are still
performing quite competitive on this data compared with other methods. Basically, this
dataset is quite small that it limits the capacity of our methods.

In BlogCatalog$^1$ and BlogCatalog$^2$, we found both GAT and GeniePath$^*$ work the
best compared with other methods. Another suprisingly interesting result is that
graph convolutional networks-style approaches can perform well even without explicit features.
This works only in transductive setting, and the look-up embeddings (in this
particular case) of testing nodes can be propagated to nodes in training.


The graph in Alipay~\cite{Liu:2017:PNN:3133956.3138827} is relative sparse.
We found that GeniePath$^*$ works quite promising on this large graph.
Since the dataset consists of ten thousands of subgraphs, the
node2vec is not applicable in this case.

We report the comparison results of inductive settings on PPI in
Table~\ref{tb:baseline_inductive}.
We use ``GCN-mean'' by averaging the neighborhoods instead of GCN for comparison.
GeniePath$^*$ performs extremely promising results
on this big graph, and shows that the adaptive depth
function plays way important compared to
GAT with only adaptive breadth function.

To further study the contributions of skip connection architectures, 
we compare GAT, GeniePath, and GeniePath-lazy 
with additional residual architecture (``skip connections'') in
Table~\ref{tb:ablation}. The results on PPI show that GeniePath is less sensitive
to the additional ``skip connections''. The significant improvement of GAT on this dataset relies
heavily on the residual structure.

In most cases, we found GeniePath-lazy converges faster than GeniePath and other GCN-style models.


\subsection{Depths of Propagation}

We show our results on classification measures with respect to the depths of propagation layers
in Figure~\ref{fig:prop_depth}. As we stack more graph convolutional layers, i.e. with deeper and broader receptive fields, 
GCN, GAT and even GraphSAGE with residual architectures can hardly maintain consistently resonable results.
Interestingly, GeniePath with adaptive path layers can adaptively learn the receptive paths and achieve
consistent results, which is remarkable.

\subsection{Qualitative Analysis}

We show a qualitative analysis about the receptive paths with respect to a sampled node learned by
GeniePath in Figure~\ref{fig:qualitative}. It can be seen, the graph Laplacian assigns importance of neighbors at
nearly the same scale, i.e. results into very dense paths (every neighbor is important). However, the
GeniePath can help select significantly important paths (red ones) to propagate while ignoring the rest, i.e.
results into much sparser paths.
Such ``neighbor selection'' processes essentially lead the direction of the receptive paths and improve
the effectiveness of propagation.


%

\section{Conclusion}
In this paper, we studied the problems of graph
convolutional networks on identifying meaningful receptive paths.
We proposed adaptive path layers with adaptive breadth and depth
functions to guide the receptive paths.
Our experiments on large benchmark data show that GeniePath
significantly outperfoms state-of-the-art approaches, and are
less sensitive to the depths of the stacked layers, or extent of neighborhood
set by hand. The success of GeniePath shows that selecting appropriate receptive
paths for different nodes is important.
In future, we expect to further study the sparsification of receptive paths, and help explain the
potential propagations behind graph neural networks in specific applications.
In addition, studying temporal graphs that the ordering of neighbors
matters could be another challenging problem.

\bibliographystyle{aaai}

\end{document}